\documentclass[10pt,twocolumn,letterpaper]{article}

\usepackage[pagenumbers]{cvpr} % To force page numbers, e.g. for an arXiv version

\usepackage[pagebackref,breaklinks,colorlinks,allcolors=cvprblue]{hyperref}
\usepackage{tikz}
\usepackage{adjustbox}
\usepackage{multirow}
\usepackage{graphicx}
\usepackage{booktabs}
\usepackage{float}
\usepackage{colortbl}
\usepackage{amsmath}

\definecolor{datasetcolor}{rgb}{0.0,0.5,0.5} % teal tone, adjust as needed
\definecolor{cvprblue}{rgb}{0.21,0.49,0.74}
\definecolor{firstb}{HTML}{FF6767}
\definecolor{thirdb}{HTML}{3564FF}
\definecolor{fovgreen}{HTML}{BED8CB}

\def\paperID{11889}
\def\confName{CVPR}
\def\confYear{2026}

\vspace{-124pt}
\title{Task-Driven Implicit Representations for \\ Automated Design of LiDAR Systems}

\newif\ifshowcomments
\showcommentstrue % Set to \showcommentsfalse to hide comments
\ifshowcomments
    \newcommand{\comment}[1]{\textcolor{olive}{{\em #1}}}
    \newenvironment{multilinecomment}[1]{\begingroup\color{olive}#1}{\endgroup}
    \newcommand{\AD}[1]{\textcolor{purple}{{\em {\bf dr.dave:} #1}}}    
    \newcommand{\nik}[1]{\textcolor{magenta}{{\em {\bf nik:} #1}}}   
    \newcommand{\sid}[1]{\textcolor{blue}{{\em {\bf sid:} #1}}}   
\else
    \newcommand{\comment}[1]{}
    
    \newcommand{\AD}[1]{}
    \newcommand{\nik}[1]{}
    \newcommand{\sid}[1]{}
    
\fi

\newenvironment{tightquation}
  {\vspace{-12pt}\begin{equation}}
  {\end{equation}\vspace{-10pt}}

\newenvironment{tight_itemize}{
\begin{itemize}[leftmargin=15pt,nosep]
  \setlength{\topsep}{0pt}
  \setlength{\itemsep}{4pt}
  \setlength{\parskip}{0pt}
  \setlength{\parsep}{0pt}
}{\end{itemize}}

\author{
    Nikhil Behari, 
    Aaron Young,
    Tzofi Klinghoffer, 
    Akshat Dave,
    Ramesh Raskar\\
    Massachusetts Institute of Technology \\
    {\tt\small \{nbehari,aryoung,tzofi,ad74,raskar\}@mit.edu}
}
\begin{document}

\twocolumn[{%
\renewcommand\twocolumn[1][]{#1}%
\maketitle
\vspace{-10mm}
\begin{center}
    \centering
    \captionsetup{type=figure}
    \includegraphics[width=\linewidth]{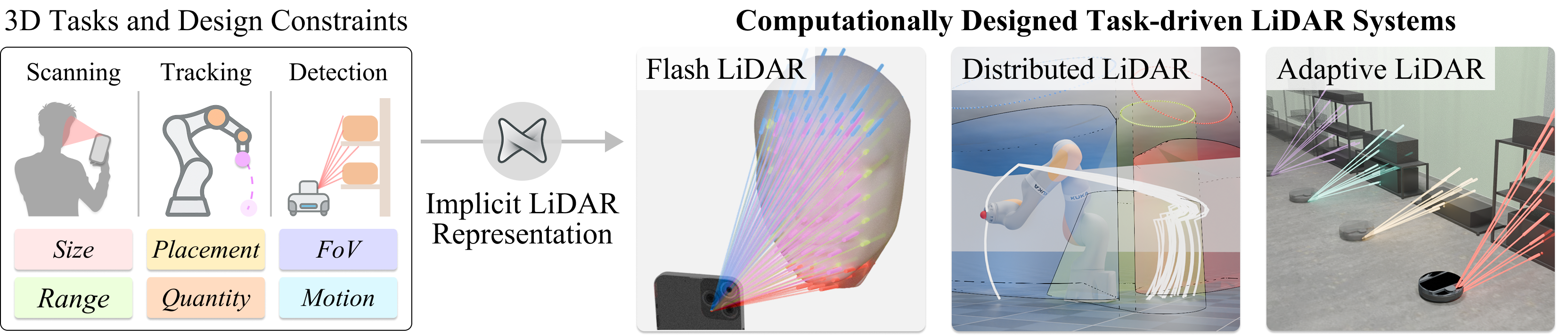}
    \vspace{-5mm}
    \captionof{figure}{\textbf{We propose a task-driven implicit representation for automated LiDAR system design.} Our implicit representation can be used to automatically propose well-suited LiDAR systems for a broad range of 3D vision tasks, such as scanning, tracking, and object detection, and can incorporate realistic physical constraints for computational LiDAR system generation.}
    \label{fig:teaser}
\end{center}%
}]

\begin{abstract}
Imaging system design is a complex and largely manual process; LiDAR design, ubiquitous in mobile, robotic, and aerial imaging, adds further complexity through unique spatial and temporal sampling requirements. In this work, we propose a framework for automated, task-driven LiDAR system design under arbitrary constraints. To achieve this, we represent LiDAR configurations in a continuous six-dimensional design space and learn task-specific implicit densities in this space via flow-based generative modeling. We then synthesize new LiDAR systems by modeling sensors as parametric distributions in 6D space and fitting these distributions to our learned implicit density using expectation-maximization, enabling efficient, constraint-aware LiDAR system design. We validate our method on diverse tasks in 3D vision, demonstrating automated LiDAR system design across real-world-inspired applications in face scanning, robotic tracking, and object detection.
\end{abstract}

% ================ INTRO 
\vspace{-5mm}
\section{Introduction} \label{sec:sec1} 
Imaging system design is a complex, time-consuming process involving the selection, placement, and optimization of optics, sensors, and illumination components. This complexity persists throughout research and development (R\&D) pipelines for mobile devices \cite{blahnik2021smartphone}, autonomous vehicles (AVs) \cite{li2020lidar}, airborne imaging platforms \cite{jia2025airborne}, and space instruments \cite{mouroulis2018review}. Current design practices rely largely on iterative manual prototyping, balancing task objectives with constraints such as manufacturing limits, illumination standards, placement restrictions, and scene-specific factors like occlusions. These challenges motivate an \textit{automated} imaging system design paradigm, which could accelerate R\&D, incorporate practical design constraints, and even discover unconventional task-optimized designs. 

We focus specifically on direct time-of-flight (dToF) LiDAR (hereafter \textit{LiDAR}) system design, which enables 3D perception in AVs, robotic imaging, and consumer devices. LiDARs infer metric depth by emitting laser pulses and measuring travel time of photon returns; they are particularly useful in low-light, textureless, or cluttered scenes where RGB perception fails. While many conventional imaging design factors apply to LiDAR systems (e.g., field of view, resolution, placement), LiDARs also introduce unique new design parameters, including scanning patterns, temporal detection windows (time gates), emitter power, and data-throughput requirements. 

LiDAR system design presents three key technical challenges. First, it requires navigating a high-dimensional, mixed discrete-continuous space encompassing sensor count, scanning patterns, placement, orientation, field of view, and temporal parameters (e.g., time gates). Second, configurations must be well-suited for the target application: a smartphone LiDAR should capture fine facial geometry, while a distributed robotic tracking system should conform to the robot's workspace and kinematic constraints. Third, designs must satisfy physical constraints (e.g., size, weight, power, range) \textit{and} user-specified design preferences -- requiring rapid, constraint-aware design exploration. 

In this work, we propose a framework for computational design of \textit{task-driven} LiDAR systems under arbitrary constraints (\cref{fig:teaser}). First, we show that LiDAR systems can be represented in a continuous six-dimensional design space (\cref{sec:sec3}). We then pose the problem of LiDAR system design as that of learning task-specific implicit densities in this 6D space (\cref{sec:sec4}). To learn, represent, and sample from this density, we leverage flow-based generative modeling to optimize a LiDAR-specific density objective under task simulations, thereby encoding task-specific LiDAR designs as \textit{implicit transformations} of simple base densities into our target design distribution (\cref{sec:sec4_3}). We then synthesize new LiDAR systems by modeling LiDAR sensors as parametric distributions in 6D space and fitting these distributions to our task-driven implicit density using expectation-maximization (\cref{sec:sec5}). As a result, our approach enables rapid, constraint-aware, task-driven LiDAR system design.

Our work draws inspiration from recent advances in implicit neural representations (INRs) \cite{sitzmann2019scene}, such as neural radiance fields (NeRFs) \cite{mildenhall2021nerf}, and demonstrates a path toward translating the core paradigm of INRs to a new class of problems in computational LiDAR design. Analogous to NeRFs, which learn an implicit volumetric function over a continuous 5D space, we propose learning an implicit density over a continuous 6D LiDAR design space. Similarly, as NeRFs enable novel-view synthesis through strategic sampling of the implicit \textit{scene representation}, we propose a constraint-aware sampling strategy from our task-driven implicit \textit{design representation} to generate physically realizable LiDAR systems, demonstrating the potential of INRs in computational sensor-system design.

We evaluate our framework across diverse 3D perception tasks: 3D face mesh reconstruction, robotic end-effector tracking, and warehouse object detection. In each case, we demonstrate that our method produces LiDAR configurations that (i) adapt to variations in scene geometry, occlusions, and motion, (ii) represent arbitrary form factors, including fixed, bi-static, distributed, and movable sensor configurations, and (iii) conform to arbitrary physical and user-imposed constraints on sensor count, placement, motion, field of view, timing, and data bandwidth, highlighting the flexibility of our approach.

\vspace{4pt}
\textbf{In summary, our key contributions are as follows:}
\begin{tight_itemize}
    \item We introduce the \textit{first continuous, implicit representation} for task-driven LiDAR system design.
    \item We represent LiDAR systems in a continuous \textit{6D design space}, and show how task-driven designs can be learned as \textit{implicit densities} over this 6D space via flow-based generative modeling. 
    \item We then \textit{synthesize physically realizable} LiDAR systems from the learned implicit density by fitting LiDAR sensors as \textit{parametric distributions} via expectation-maximization.
    \item  We validate our framework on \textit{diverse 3D perception tasks}, including 3D reconstruction, object tracking, and object detection in scenarios inspired by applications in mobile sensing and robotics.
\end{tight_itemize}

\vspace{-1mm}
\section{Related Work}
\vspace{-1mm}
\textbf{LiDAR Design Diversity.} LiDAR has become a pervasive sensing modality used in AVs \cite{li2020lidar,ignatious2022overview}, smartphones \cite{rangwala2020iphone12lidar}, and robotics \cite{yang20223d,khan2021comparative}. Recent hardware advancements have enabled production of low-cost consumer LiDARs: single-pixel (e.g. ST VL53L0X) and multi-pixel (e.g. STVL53L5CX, AMS TMF8828) dToF sensors have been used in assistive robotics \cite{kramer2023healthwalk,ghafoori2024novel}, healthcare IoT \cite{leng2022design}, robotic localization \cite{caroleo2025tiny}, autonomous navigation \cite{young2024enhancing}, and 3D scene reconstruction \cite{mu2024towards,behari2024blurred}. 

\noindent \textbf{Sensor Placement and Automated Design.} Next-best-view methods optimize 3D reconstruction fidelity via rule- or learning-based viewpoint selection \cite{guedon2022scone,pan2022activenerf,peralta2020next,ran2023neurar,chaplot2020learning,hamdi2021mvtn,chen2024gennbv}, but assume \textit{fixed camera hardware} and tasks. Similarly, LiDAR optimization methods for AVs \cite{li2024your,liu2019should,hu2022investigating,li2024influence} focus only on the \textit{placement} of \textit{scanning-based} LiDARs. In contrast, we model the continuous LiDAR design space, spanning flash, gated, and motion-adaptive LiDARs. Recent joint camera–perception design methods \cite{klinghoffer2023diser, yan2025tacos} tune predefined camera parameters for fixed objectives, requiring retraining when constraints change. We instead propose a new \textit{representation} for LiDAR systems that supports arbitrary post-hoc constraint adjustment without retraining.

\begin{figure*}[th!]
\centering
\includegraphics[width=\linewidth]{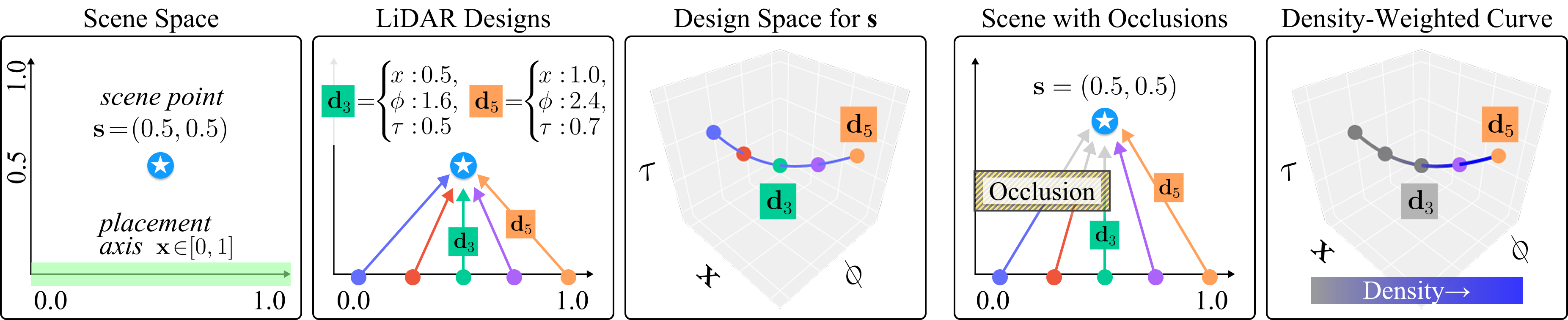}
\vspace{-6mm}
\captionof{figure}{\textbf{Visualization of simplified (6D$\rightarrow$3D) LiDAR design space for a single 2D scene point.} Scene points can be observed by infinitely many geometrically-equivalent LiDAR designs, which form density-weighted curves in our design space. We then further weight design points along curves by their ray visibility, incorporating scene occlusions.\vspace{-2.5mm}}
\label{fig:fig2a_representation}
\vspace{-4pt}
\end{figure*}

\noindent \textbf{Neural Implicit Representations.} Our work draws on recent paradigms in implicit neural representations (INRs), such as neural radiance fields (NeRFs) \cite{mildenhall2021nerf}, neural implicit surface representations \cite{park2019deepsdf,yariv2021volume}, and implicit occupancy networks \cite{mescheder2019occupancy}. In general, research in INRs \cite{sitzmann2020implicit,martel2021acorn,lindell2022bacon,saragadam2023wire} has primarily focused on representing \textit{visual} signals; in this work, we propose extending INRs to represent the \textit{design of LiDAR imaging hardware} itself.

% 6D lidar representation. 

\vspace{-2mm}
\section{Representing LiDARs in 6D Design Space} \label{sec:sec3}
\vspace{-1mm}
In dToF LiDAR, a pulsed laser emitter emits short optical bursts along a ray direction. These pulses reflect off scene surfaces and are detected by a co-located receiver. The elapsed time $\tau$ between emission and detection (the ``time of flight") directly encodes scene depth along the ray, computed as $\textrm{depth} = (c\tau)/2$, where $c$ is the speed of light. In this work, we describe the design of \textit{LiDAR systems}, which consist of multiple LiDAR sensors; each sensor is a collection of these time-of-flight (ToF) measurements, each taken at a discrete spatial origin, angular direction, and time-of-flight detection range. 

\vspace{0.2mm}
\noindent \textbf{6D LiDAR Parameterization.} Every LiDAR measurement can be represented in a \textit{6D design space}, where each point represents a ray with an infinitesimal measured ToF distance. Specifically, a LiDAR measurement is defined by its spatial origin $\mathbf{x} = (x, y, z) \in \mathbb{R}^3$, ray direction (azimuth $\phi$ and elevation $\psi$) angle $\mathbf{a} = (\phi, \psi) \in [0,2\pi) \times \left[-\frac{\pi}{2}, \frac{\pi}{2}\right]$, and temporal coordinate $\tau \in \mathbb{R}^+$ representing ToF (or equivalently, depth) ``observed" along the ray. Thus, every LiDAR measurement can be expressed as a \textit{design point}:

\vspace{1.5mm}
\begin{tightquation}
\mathbf{d} = (\mathbf{x}, \mathbf{a}, \tau)^\top = (x, y, z, \phi, \psi, \tau)^\top \in \mathcal{D}
\end{tightquation}
\vspace{-1.5mm}

\noindent in 6D LiDAR design space $\mathcal{D} = \mathcal{X}\times\mathcal{A}\times\mathcal{T}$. That is, each point $\mathbf{d} \in \mathcal{D}$ encodes a LiDAR measurement at a single origin, angle, and time. \cref{fig:fig2a_representation} illustrates this design space for LiDARs in a simplified flatland scene. 

A complete LiDAR return can then be represented by pairing an emitter ray $\mathbf{d}_e = (\mathbf{x}_e, \mathbf{a}_e, \tau_e)$ and detector ray $\mathbf{d}_d = (\mathbf{x}_d, \mathbf{a}_d, \tau_d)$. The emitter at spatial origin $\mathbf{x}_e$ sends a Dirac delta pulse along direction $\mathbf{a}_e$, which reflects at a scene point $\mathbf{s}$ and is detected at $\mathbf{x}_d$. Terms $\tau_e$ and $\tau_d$ then denote one-way travel times from emitter to surface and from surface to detector, giving a total measured ToF of $\tau_e + \tau_d$. By Helmholtz reciprocity \cite{helmholtz1925treatise}, swapping emitter and detector does not affect the return. Thus, measurable LiDAR returns can be described as tuples \((\mathbf{d}_e, \mathbf{d}_d)\) in our 6D space which satisfy $\mathcal{M}(\mathbf{d}_e) = \mathcal{M}(\mathbf{d}_d) = \mathbf{s}$, where \(\mathcal{M}: \mathcal{D} \rightarrow \mathcal{S} \subset \mathbb{R}^3\) maps each LiDAR ray $\mathbf{d}$ to its observed scene point $\mathbf{s}$. Concretely, for $\mathbf{d} = (\mathbf{x}, \mathbf{a}, \tau)^\top$, 

% \begin{tightquation}
% \mathbf{s} = \mathcal{M}(\mathbf{d})
%   = \mathbf{x} + \tau\,\mathbf{v}(\phi,\psi),\quad 
% \mathbf{v}(\phi,\psi)
%   = \bigl(\cos\psi\cos\phi,\;\cos\psi\sin\phi,\;\sin\psi\bigr)^{\!\top}.
% \end{tightquation}

\begin{tightquation}
\begin{aligned}
\mathbf{s}
&= \mathcal{M}(\mathbf{d})
 = \mathbf{x} + \tau\,\mathbf{v}(\phi,\psi), \\[-2pt]
\mathbf{v}(\phi,\psi)
&= \bigl(
    \cos\psi\cos\phi,\;
    \cos\psi\sin\phi,\;
    \sin\psi
  \bigr)^{\!\top}.
\end{aligned}
\end{tightquation}
\vspace{1mm}

\begin{figure*}[t]
\centering
\includegraphics[width=\linewidth]{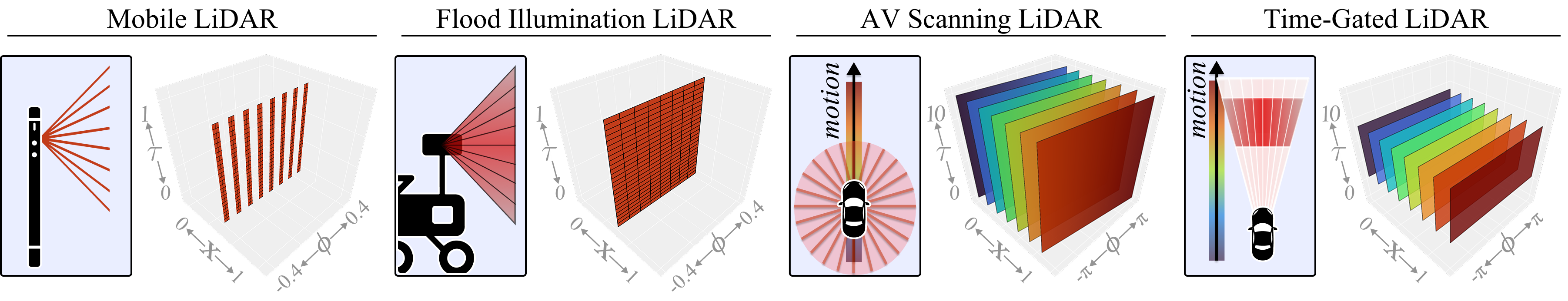}
\captionof{figure}{\textbf{Existing LiDARs represented in design space.} LiDARs form discrete sampling volumes in our 6D design space, reflecting sensor origin, motion, direction, FoV, and time gating.\vspace{-2.5mm}}
\label{fig:fig3a_lidarsindesign}
\vspace{-6pt}
\end{figure*}

\vspace{0.2mm}
\noindent \textbf{Representing Existing LiDARs in 6D Space}. Within this continuous space, \textit{all} LiDARs can be viewed as sampling \textit{discrete volumes} of LiDAR origins $\{\mathbf{x}_j\}\subset\mathcal{X}$, directions $\{\mathbf{a}_k\}\subset\mathcal{A}$, and times $\{\tau_\ell\}\subset\mathcal{T}$. To demonstrate this, we visualize several existing LiDAR designs within a simplified (6D $\rightarrow$ 3D) design space in \cref{fig:fig3a_lidarsindesign}. In this work, we focus on co-located emitter-detector LiDARs where $\mathbf{d}_e = \mathbf{d}_d$, and thus each sampled design point represents both an emitter and corresponding detector ray (although our framework also supports bi-static LiDAR design, shown \cref{fig:figure5}). 

% BISTATIC Des.. While co-located design is common for LiDARs, measurable LiDAR returns only strictly require the intersection of emitter and detector rays at a surface point, a principle leveraged in bistatic LiDAR systems \cite{tobin2021robust,scholes2024robust}. Our framework naturally handles both design cases by enforcing $\mathcal{M}(\mathbf{d}_e) = \mathcal{M}(\mathbf{d}_d) = \mathbf{s}$ so every sampled detector $\mathbf{d}_d$ has a corresponding emitter $\mathbf{d}_e$ mapping to the same scene point $\mathbf{s}$. 

 \vspace{-2mm}
\section{Learning Task-Specific LiDAR Designs as Implicit Densities} \label{sec:sec4}
\vspace{-1mm}
3D computer vision tasks (e.g., 3D scanning, tracking, detection) are generally defined over distributions of scenes that capture task-relevant variability. For example, face scanning may try to capture fine detail across facial geometries, whereas a robot tracking method may consider a distribution of kinematically constrained trajectories across frames. Our objective, leveraging scene simulations for a given task, is to \textit{project this scene variability into our LiDAR design space}, to learn the subspace of high-quality LiDAR designs for a given task. 

To represent this subspace of high-quality designs, we propose learning a \textit{task-driven implicit density} over the 6D LiDAR design space, which concentrates on high-quality designs for a class of scenes (i.e., a task). Specifically, we propose learning an implicit density $p_{\text{task}}(\mathbf{d}; \Theta) : \mathcal{D} \rightarrow \mathbb{R}^{+}$ where high-density regions correspond to well-suited LiDAR configurations for a given task. However, to learn this continuous density, we must first define a metric that quantifies the \textit{quality} of each design point. This \textit{target density} $p^*(\mathbf{d})$ must (1) \textit{generalize} across arbitrary scene types and tasks, and (2) reflect the \textit{physical properties} of valid LiDAR returns to accurately project scenes into LiDAR designs.

\subsection{Density Evaluation for LiDAR Measurements. } The core objective of LiDAR design is to maximize valid returns from target surfaces, thereby improving 3D understanding--notably, this goal generalizes across tasks including geometry reconstruction, motion estimation, and object detection. We therefore propose a target density formulation that expresses this objective over the 6D LiDAR design space. To do this, we weight the density of each candidate ray $\mathbf{d}$ by two factors. First, design points $\mathbf{d}$ must have \textit{high surface proximity}: for a class of scenes, the scene point $\mathbf{s} = \mathcal{M}(\mathbf{d})$ captured by LiDAR measurement $\mathbf{d}$ should, on average, lie on or near \textit{a surface} to maximize 3D understanding. Second, design points should have \textit{high visibility}: the ray path for measurement $\mathbf{d}$ from origin $\mathbf{x}$ to scene point $\mathbf{s}$ must have high transmittance, i.e., be free of occlusions. 

To compute this density, we consider a class of task-specific scenes (e.g, face meshes for face scanning) represented by signed distance functions (SDFs). For each design point $\mathbf{d}$, we evaluate the two density terms (surface proximity and ray visibility) \textit{over the class of scenes}. Let $\mathrm{SDF}_i(\mathbf{s})$ be the signed distance, for scene $i$, from scene point $\mathbf{s}$ to the nearest surface (zero at the surface, positive outside). We then define an energy function measuring the \textit{surface proximity} of each design point $\mathbf{d}$ with observed scene point $\mathbf{s} = \mathcal{M}(\mathbf{d})$: 

\vspace{1.5mm}
{\small
\begin{tightquation}
\mathrm{S}_i(\mathbf{d}) = \exp\!\left(-\frac{\mathrm{SDF}_i(\mathbf{s})^2}{2\sigma^2}\right),
\end{tightquation}
}
\vspace{1.5mm}

\noindent which describes the likelihood, under Gaussian noise $\sigma$, that a scene point $\mathbf{s}$ observed by design $\mathbf{d}$ at time $\tau$ lies near a surface in scene $i$. We then model the \textit{visibility} of ray $\mathbf{d}$ in scene $i$, from its origin $\mathbf{x}$ to scene point $\mathbf{s}$, using a transmittance formulation from volumetric rendering \cite{max2002optical}: 

\vspace{-0.5mm}
% {\footnotesize
% \begin{tightquation}
% \mathrm{T}_i(\mathbf{d}) = \exp\!\left(-\int_{0}^{\tau} \beta\, \cdot\text{sigmoid}\bigl(-\mathrm{SDF}_i\bigl(\mathbf{x} + \lambda\, \mathbf{v}(\phi,\psi)\bigr)\bigr)\, d\lambda\right),
% \end{tightquation}
% }
{\small
\begin{tightquation}
\begin{split}
\mathrm{T}_i(\mathbf{d}) &= \\
&\hspace{-10pt}\exp\!\left(
  -\!\int_{0}^{\tau} 
    \beta \,
    \text{sigmoid}\!\left(
      -\mathrm{SDF}_i\!\left(\mathbf{x} + \lambda\, \mathbf{v}(\phi,\psi)\right)
    \right)
  d\lambda
\right)
\end{split}
\end{tightquation}
}

\vspace{0.5mm}

\noindent where \(\beta > 0\) scales attenuation. This term quantifies the \textit{visibility} of a design point across a class of scenes. Our final target density is then defined as the combined surface-proximity and visibility terms for each design point $\mathbf{d}$ over the sum of scenes $i$: 

\vspace{1mm}
\begin{tightquation}
p^*(\mathbf{d}) = \sum_i
  \underbrace{\mathrm{S}_i(\mathbf{d})}_{\substack{\text{Surface}\\\text{Proximity}}}
  \, \times 
  \underbrace{\mathrm{T}_i(\mathbf{d})}_{\substack{\text{Ray}\\\text{Visibility}}}.
\end{tightquation}

% ==============

\begin{figure*}[t]
\centering
\includegraphics[width=0.95\linewidth]{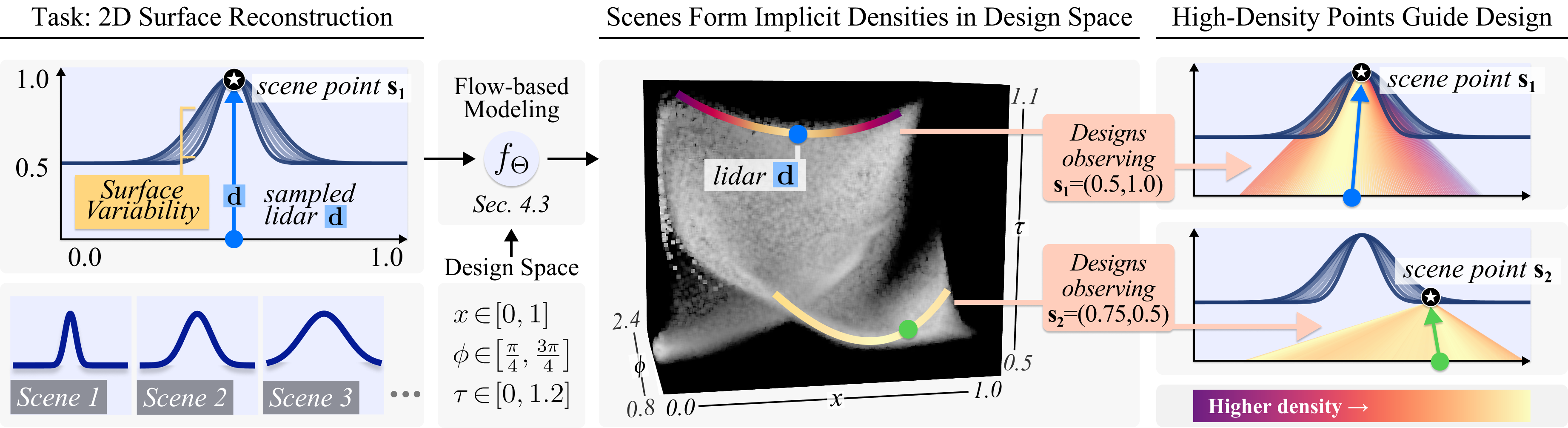}
% \framebox[\linewidth]{\rule{0pt}{2in}}
\captionof{figure}{\textbf{Visualization of implicit LiDAR density for 2D surface scenes in simplified (6D$\rightarrow$3D) design space.} We visualize a learned implicit LiDAR density over the parameter domain $(x \in [0,1], \phi \in [\frac{\pi}{4},\frac{3\pi}{4}],\tau\in[0,1.2])$ for a class of 2D scenes. High-density points in this space, representing high-proximity and high-visibility measurements, can be used to guide LiDAR sensor design.}
\label{fig:fig2b_designspacesurfaces}
\vspace{-8pt}
\end{figure*}

\vspace{-2mm}
\subsection{Projecting Tasks into LiDAR Design Space.} We interpret the learned design distribution under our target density as a projection of 3D task variability into the LiDAR design space. The \textit{surface proximity} term encodes the frequency of surface points across scenes, capturing variations in both object geometry and motion. Given a fixed scene point $\mathbf{s}$ with full visibility $\mathrm{T}(\mathbf{d}) = 1$ for design points $d \in \mathcal{M}^{-1}(\mathbf{s})$, if that scene point $\mathbf{s}$ lies on the surface in $I$ scenes, then we have target density $p^*(\mathbf{d})=\sum_i\exp(-\mathrm{SDF}_i(\mathbf{s})^2/(2\sigma^2))\approx\sum_{i\in I}1=|I|$, proportional to the frequency of surface point $\mathbf{s}$ across the scene class. Similarly, the \textit{visibility} term distinguishes LiDAR designs by incorporating occlusions through ray transmittance weighting. Together, these terms help us project task-specific scene variability into high surface-proximity and high-visibility LiDAR designs. We visualize this projection in a simplified (6D$\rightarrow$3D) design space for a single scene point in \cref{fig:fig2a_representation} and for a class of scenes in \cref{fig:fig2b_designspacesurfaces}. 

We note that the surface proximity term introduces ambiguity in the design space that reflects the inherent ambiguity of real LiDAR design. For a given scene point $\mathbf{s}$, infinitely many LiDAR designs can perfectly ``observe'' that scene point under different origins, angles, and temporal configurations. The forward mapping $\mathcal{M}$ captures this many-to-one correspondence of geometrically-equivalent rays as the set $\mathcal{M}^{-1}(\mathbf{s}) = \{\mathbf{d} \in \mathcal{D} \mid \mathcal{M}(\mathbf{d}) = \mathbf{s}\}$. Importantly, our surface proximity term models this geometric equivalence, allowing us to visualize rays that observe the same point $\mathbf{s}$ as \textit{curves} in design space. We note that our ray visibility term further weights equivalent designs by their transmittance, yielding a physically grounded projection of scene geometry into LiDAR design space. We illustrate this in a simplified (6D$\rightarrow$3D) design space in \cref{fig:fig2a_representation} and \cref{fig:fig2b_designspacesurfaces}.

% =====================

\vspace{-2mm}
\subsection{Learning Continuous LiDAR Design Densities via Flow-Based Modeling} \label{sec:sec4_3}

Although we have a well-defined evaluation metric for each design point, directly identifying high-density regions for a given task remains intractable. Our key insight is to use flow-based generative modeling to efficiently \textit{learn} and \textit{sample from} task-specific target distributions. By learning a mapping from an easy-to-sample base distribution to our complex target density, we thereby represent each LiDAR design density as an \textit{implicit transformation} of the tractable base. To learn this transformation, we leverage normalizing flows, a technique widely used in generative modeling \cite{papamakarios2021normalizing} for efficiently learning transformations between distributions; these learned transformations are invertible and differentiable, yielding closed-form Jacobian determinants and exact likelihood computation. 

To leverage normalizing flows for learning this task-driven implicit density, we begin with latent samples $\mathbf{z}\sim\pi$ drawn from a 6D uniform \textit{base distribution} $\pi = U\bigl([0,1]^6\bigr)$. We then learn an invertible mapping \(f : \mathbb{R}^6 \to \mathbb{R}^6\) that transforms \(\mathbf{z}\) into a design point \(\mathbf{d} = f(\mathbf{z})\). The resulting density is obtained via the change‐of‐variables formula:

\vspace{2.5mm}
{\small
\begin{tightquation}
p(\mathbf{d}) = \pi\!\Bigl(f^{-1}(\mathbf{d})\Bigr)\,\Bigl|\det\!\Bigl(\nabla f^{-1}(\mathbf{d})\Bigr)\Bigr|.
\end{tightquation} 
}
\vspace{1mm}

% We realize our flow from $K$ autoregressive layers \cite{huang2018neural}; at layer $i$ we update one coordinate by $d_i=h_{\Theta_i}(z_i;z_{1:i-1})$, where an MLP that sees the preceding latents $z_{1:i-1}$ provides the parameters $\Theta_i$.  This causal structure makes the Jacobian strictly lower‑triangular, so we evaluate $\log|\det\nabla\mathbf f|$ with an $O(D)$ sum over its diagonal entries.  To implement each one‑dimensional transform $h_{\Theta_i}$ we use a monotone rational‑quadratic neural spline flow \cite{durkan2019neural}, which can approximate complex monotone warps with closed‑form inverses and Jacobian determinants. Before applying the spline we rescale the uniform base variables $z_i\!\in[0,1]$ into the spline’s domain, and the MLP supplies the knots’ widths, heights, and end‑slopes that fully specify the transformation.
 
In practice, the mapping \(f\) is learned as a composition of invertible functions, \(f = f_K \circ \cdots \circ f_1\). We implement $f$ using $K$ autoregressive spline-flow layers \cite{huang2018neural,durkan2019neural}, where each coordinate update $d_i = h_{\Theta_i}(z_i; z_{1:i-1})$ uses an MLP-conditioned rational-quadratic spline whose knot widths, heights, and end-slopes are predicted by the MLP. During training, we rescale uniform base samples $z_i \in [0,1]$ into the spline’s input domain, apply the spline transform, and then rescale the outputs to the 6D LiDAR design space. We train our flow model by minimizing reverse Kullback–Leibler (KL) divergence between the learned density \(p(\mathbf{d};\Theta)\) and the target density \(p^*(\mathbf{d})\), where \(p^*(\mathbf{d})\) is estimated from \textit{task-specific scene simulations}. The final loss is then given by

% {
% \footnotesize 
% \begin{tightquation}
% \begin{aligned}
% \mathcal{L}(\Theta) &= \mathbb{E}_{\mathbf z\sim\pi} 
% \Biggl[
%   \underbrace{(1+\eta_{\text{entropy}})\vphantom{\Bigl(\log \pi(\mathbf{z}) - \log\Bigl|\det\!\Bigl(\nabla f(\mathbf{z})\Bigr)\Bigr| \Bigr)}}_{\text{Entropy Reg.}} \, 
%   \underbrace{\Bigl(\log \pi(\mathbf{z}) - \log\Bigl|\det\!\Bigl(\nabla f(\mathbf{z})\Bigr)\Bigr| \Bigr)}_{\substack{\text{Flow-based Log Density}}}
%   \quad - \underbrace{\log p^*\bigl(f(\mathbf{z})\bigr)\vphantom{\Bigl(\log \pi(\mathbf{z}) - \log\Bigl|\det\!\Bigl(\nabla f(\mathbf{z})\Bigr)\Bigr| \Bigr)}}_{\substack{\text{LiDAR Target Log Density}}}
% \Biggr].
% \end{aligned}
% \end{tightquation}
% }

{\small
\begin{tightquation}
\begin{aligned}
\mathcal{L}(\Theta)
&= \mathbb{E}_{\mathbf z\sim\pi}
\Biggl[
  \underbrace{(1+\eta_{\text{ent}})%
  \vphantom{\Bigl(\log \pi(\mathbf{z}) - 
  \log\Bigl|\det\!\Bigl(\nabla f(\mathbf{z})\Bigr)\Bigr| \Bigr)}}_{\text{Entropy Reg.}}
  \,
  \underbrace{\Bigl(
    \log \pi(\mathbf{z})
    - \log\Bigl|\det\!\Bigl(\nabla f(\mathbf{z})\Bigr)\Bigr|
  \Bigr)}_{\substack{\text{Flow-based}\\\text{Log Density}}}
  \\[-4pt]
  &\hspace{20pt}\qquad
  -\,\underbrace{\log p^*\bigl(f(\mathbf{z})\bigr)
  \vphantom{\Bigl(\log \pi(\mathbf{z}) - 
  \log\Bigl|\det\!\Bigl(\nabla f(\mathbf{z})\Bigr)\Bigr| \Bigr)}}_{\substack{\text{LiDAR Target}\\\text{Log Density}}}
\Biggr].
\end{aligned}
\raisetag{5pt}
\end{tightquation}
}
\vspace{1mm}

\noindent We add entropy regularization term $\eta_{\text{ent}}$ to promote sampling diversity during training. Additional implementation and training details are provided in the supplement. 

% ======================================

 \vspace{-1mm}
\section{Generating LiDAR Systems from Task-Driven Implicit Densities} \label{sec:sec5}

\begin{figure}[b]
  \vspace{-2mm}  
  \centering
  \includegraphics[width=\columnwidth]{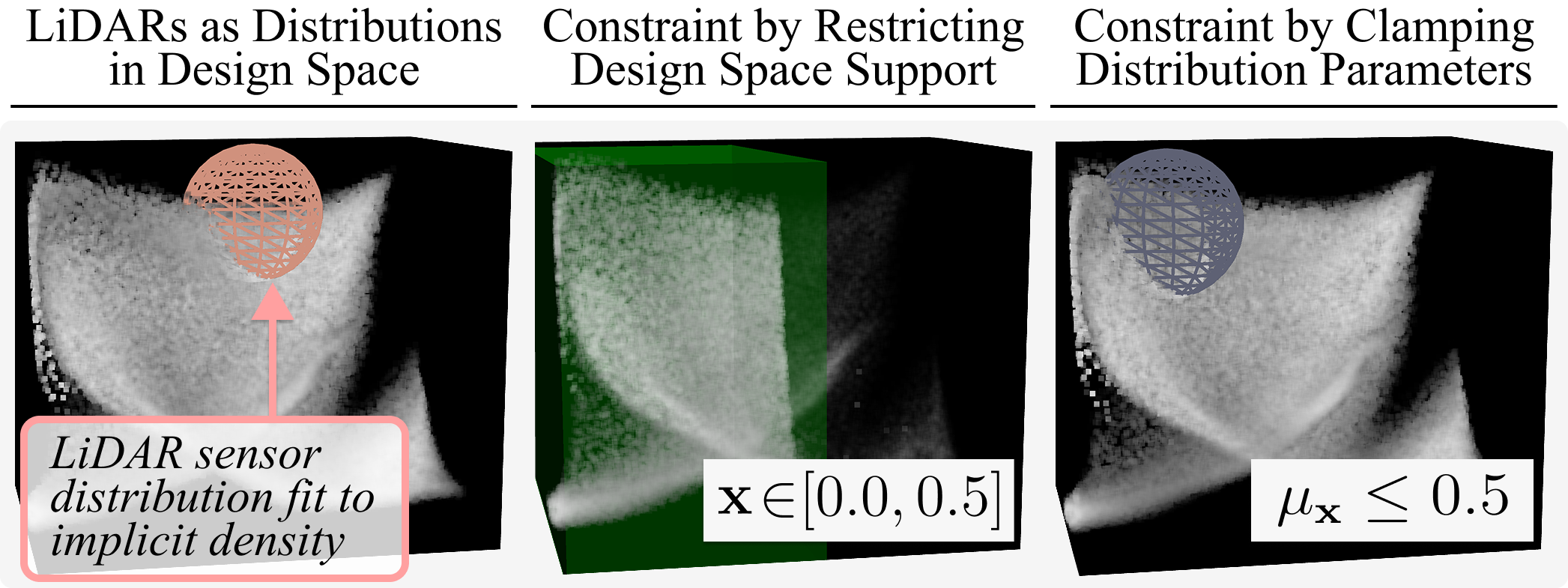}
  \vspace{-6mm}
  \captionof{figure}{\textbf{Representing Sensors and Imposing Constraints in LiDAR Design Space.} Sensors are represented as parametric distributions in 6D design space, and are learned via expectation-maximization (EM) over the task-driven implicit density. Physical or user-imposed constraints can be easily incorporated by restricting the design space support or by clamping the parameters of the learned sensor distributions.}
  \label{fig:fig3b_constraints}
\end{figure}

\textbf{Sensors as Distributions for LiDAR System Design.}
Having shown that existing LiDARs correspond to discrete sampling volumes within our continuous 6D design space, our final insight is that new LiDAR sensors and systems can be modeled as \textit{parametric probability distributions} over design space $\mathcal{D}$. Specifically, let $p(\mathbf{d})$ be the learned task-driven implicit density from \cref{sec:sec4_3}. We model a new sensor as $q(\mathbf d\mid\theta)=q_x(\mathbf x\mid\theta_x)\,q_a(\mathbf a\mid\theta_a)\,q_\tau(\tau\mid\theta_\tau)$, where $\theta = \{\theta_x,\theta_a,\theta_\tau\}$ parametrizes the distributions over spatial origin $\mathbf{x}\in\mathbb{R}^3$, angle $\mathbf{a}\in\mathbb{R}^2$, and time‐of‐flight $\tau\in\mathbb{R}^+$. This formulation allows us to represent and enforce constraints over sensor characteristics, including placement and motion, sensor FoV and beam pattern, and temporal gating. 

\begin{figure*}[th]
  \centering
  % \fbox{\parbox{\textwidth}{\centering Top Figure Placeholder}}
  \includegraphics[width=\textwidth]{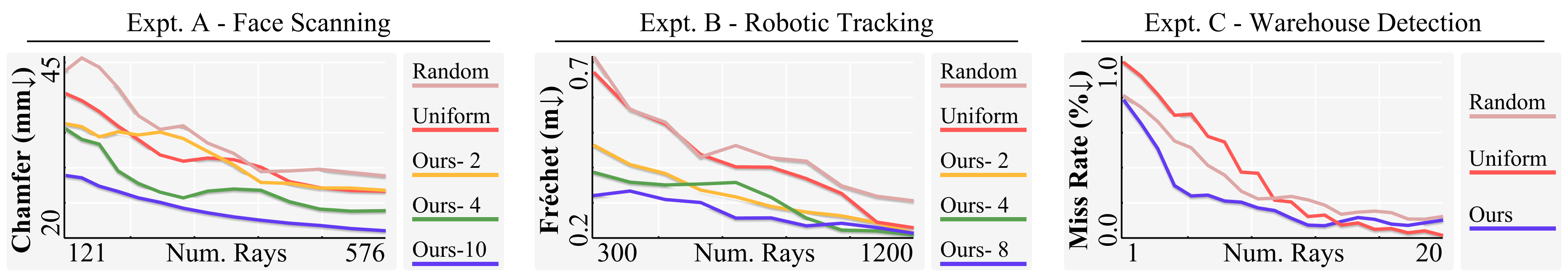}
  
  \vspace{0.2em}
  
\begin{minipage}[t]{0.32\textwidth}
  \centering
  {\scriptsize
   \renewcommand{\arraystretch}{1.2}
   \resizebox{\linewidth}{!}{%
     \begin{tabular}{l c c c}
       \hline \\[-2.75ex]
             & \multicolumn{3}{c}{Bandwidth @10Hz, 40-bit bins} \\[0.25ex]
             & 196 Rays & 361 Rays & 576 Rays \\
       \hline \\[-3ex]
       \colorbox[HTML]{F5E4E4}{BL: Random}      
                     & \shortstack{5.2 Mbps}
                     & \shortstack{9.5 Mbps}
                     & \shortstack{15.2 Mbps}   \\
       \hline \\[-3ex]
       \colorbox[HTML]{FFCCCC}{BL: Uniform}     
                     & \shortstack{5.2 Mbps}
                     & \shortstack{9.5 Mbps}
                     & \shortstack{15.2 Mbps}   \\
       \hline \\[-3ex]
       \colorbox[HTML]{FEEAC3}{Ours: 2 Sensors} 
                     & \shortstack{\textbf{0.9 Mbps}}
                     & \shortstack{\textbf{1.6 Mbps}}
                     & \shortstack{\textbf{2.5 Mbps}}    \\
       \hline \\[-3ex]
       \colorbox[HTML]{CDE2CA}{Ours: 4 Sensors} 
                     & \shortstack{\textbf{0.9 Mbps}}
                     & \shortstack{\textbf{1.7 Mbps}}
                     & \shortstack{\textbf{2.7 Mbps}}    \\
       \hline \\[-3ex]
       \colorbox[HTML]{CFC3FB}{Ours: 10 Sensors}
                     & \shortstack{\textbf{0.8 Mbps}}
                     & \shortstack{\textbf{1.5 Mbps}}
                     & \shortstack{\textbf{2.3 Mbps}}    \\
       \hline
     \end{tabular}%
   }
  }
\end{minipage}
  \hfill
    \begin{minipage}[t]{0.32\textwidth}
      \centering
      {\scriptsize                         % 1) shrink everything
       \renewcommand{\arraystretch}{1.2}   %    tighten row spacing
       \resizebox{\linewidth}{!}{%         % 2) then stretch to full width
         \begin{tabular}{l c c c}
           \hline \\[-2.75ex]
                 & \multicolumn{3}{c}{Bandwidth @10Hz, 40-bit bins} \\[0.25ex]
                 & 400 Rays & 1000 Rays & 1200 Rays \\
           \hline \\[-3ex]
           \colorbox[HTML]{F5E4E4}{BL: Random}
                         & \shortstack{10.7 Mbps}
                         & \shortstack{26.8 Mbps}
                         & \shortstack{32.2 Mbps}   \\
           \hline \\[-3ex]
           \colorbox[HTML]{FFCCCC}{BL: Uniform}
                         & \shortstack{10.7 Mbps}
                         & \shortstack{26.8 Mbps}
                         & \shortstack{32.2 Mbps}   \\
           \hline \\[-3ex]
           \colorbox[HTML]{FEEAC3}{Ours: 2 Sensors}
                         & \shortstack{\textbf{6.9 Mbps}}
                         & \shortstack{\textbf{17.3 Mbps}}
                         & \shortstack{\textbf{20.8 Mbps}}    \\
           \hline \\[-3ex]
           \colorbox[HTML]{CDE2CA}{Ours: 4 Sensors}
                         & \shortstack{\textbf{5.4 Mbps}}
                         & \shortstack{\textbf{13.4 Mbps}}
                         & \shortstack{\textbf{16.1 Mbps}}    \\
           \hline \\[-3ex]
           \colorbox[HTML]{CFC3FB}{Ours: 8 Sensors}
                         & \shortstack{\textbf{4.7 Mbps}}
                         & \shortstack{\textbf{11.7 Mbps}}
                         & \shortstack{\textbf{14.0 Mbps}}    \\
           \hline
         \end{tabular}%
       }
      }
    \end{minipage}
  \hfill
\begin{minipage}[t]{0.32\textwidth}
  \centering
  {\scriptsize
   \renewcommand{\arraystretch}{1.2}
   \resizebox{0.9\linewidth}{!}{%
     \begin{tabular}{l c c c}
       \hline \\[-2.75ex]
             & \multicolumn{3}{c}{Bandwidth @10Hz, 40-bit bins} \\[0.25ex]
             & 6 Rays & 12 Rays & 18 Rays \\
       \hline \\[-3ex]
       \colorbox[HTML]{F5E4E4}{BL: Random}
                     & \shortstack{1.9 Mbps}
                     & \shortstack{3.8 Mbps}
                     & \shortstack{5.8 Mbps}      \\
       \hline \\[-3ex]
       \colorbox[HTML]{FFCCCC}{BL: Even}
                     & \shortstack{1.9 Mbps}
                     & \shortstack{3.8 Mbps}
                     & \shortstack{5.8 Mbps}     \\
       \hline \\[-3ex]
       \colorbox[HTML]{CFC3FB}{Ours}
                     & \shortstack{\textbf{0.3 Mbps}}
                     & \shortstack{\textbf{0.4 Mbps}}
                     & \shortstack{\textbf{0.6 Mbps}}     \\
       \hline
     \end{tabular}%
   }
  }
\end{minipage}
  \vspace{-3pt}
  \caption{\textbf{Quantitative task evaluations using automatically designed LiDAR systems.} Our synthesized LiDAR systems enable consistently better performance in scanning (Chamfer distance), tracking (Fréchet distance), and detection (miss rate) across varying sensor counts and sampling constraints, while dramatically reducing required bandwidth through improved time-gating. Curves are smoothed (w=3) for visual clarity; extended results and evaluation are included in the supplement. }
  \label{fig:results_metrics}
  \vspace{-9pt}
\end{figure*}

\vspace{2mm}
\noindent \textbf{Generating Sensors via Expectation Maximization.} Representing LiDAR sensors as distributions allows us to formulate sensor synthesis as maximum-likelihood estimation (MLE): $\theta^*=\arg\max_\theta \mathbb{E}_{\mathbf{d}\sim p(\mathbf{d})}[\log q(\mathbf{d}|\theta)]$. That is, we fit the parametric sensor distribution to assign the highest likelihood to high-density regions of the learned task-driven implicit density. Posing LiDAR system design as MLE over the implicit density $p(\mathbf d)$ enables a broad class of LiDAR synthesis strategies. In this work, we model $q(\mathbf d|\theta)$ as a latent‐variable Gaussian mixture model and fit $\theta$ via expectation–maximization (EM), an iterative two‐step procedure. In the expectation (E) step, we compute the posterior $q(g|\mathbf d;\theta^{(t)})$, where $g$ indexes mixture components; in the maximization (M) step, we update 
$
\theta^{(t+1)}
= \arg\max_\theta
  \mathbb{E}_{\mathbf d\sim p(\mathbf d)}
  [
    \sum_{g=1}^G
      q\bigl(g\mid \mathbf d;\theta^{(t)}\bigr)\,
      \log q(\mathbf d,g\mid\theta)
  ].
$ EM maximizes a Jensen lower bound on the log‐likelihood:

% \begin{tightquation}
% \log q(\mathbf d\mid\theta)
% = \log\!\sum_{g=1}^G q(\mathbf d,g\mid\theta)
% \;\ge\;
% \sum_{g=1}^G q\bigl(g\mid\mathbf d;\theta^{(t)}\bigr)\,
%   \log\frac{q(\mathbf d,g\mid\theta)}{q\bigl(g\mid\mathbf d;\theta^{(t)}\bigr)}.
% \end{tightquation}

\vspace{-0.5mm}
{\small
\begin{tightquation}
\begin{aligned}
\log q(\mathbf d\mid\theta)
&= \log\!\sum_{g=1}^G q(\mathbf d,g\mid\theta) \\
&\ge\;
\sum_{g=1}^G q\bigl(g\mid\mathbf d;\theta^{(t)}\bigr)\,
  \log\frac{q(\mathbf d,g\mid\theta)}%
           {q\bigl(g\mid\mathbf d;\theta^{(t)}\bigr)}.
\end{aligned}
\end{tightquation}
}
\vspace{0.5mm}

We instantiate $q(\mathbf d|\theta)$ as a $G$-component Gaussian mixture model,  
$
q(\mathbf d|\theta)
= \sum_{g=1}^G \pi_g\,\mathcal N\bigl(\mathbf d;\mu_g,\Sigma_g\bigr),
$ where $\theta=\{\pi_g,\mu_g,\Sigma_g\}_{g=1}^G$ defines a learned configuration of $G$ sensor primitives. Each Gaussian represents a learned ``sensor'', from which physical sensor parameters are derived using the $95\%$ confidence intervals in angle, time, and (for distributed/dynamic systems) origin. When ray-based LiDAR sampling is desired, each Gaussian could, in principle, represent a single ray; in practice, we sample rays uniformly within the 95\% origin/angular/time bounds of each learned sensor. In experiments where the number of rays is constrained, rays are allocated to each sensor in proportion to its mixture weight $\pi_g$, aligning spatial sampling with learned sensor importance.

\vspace{0.5mm}
\noindent \textbf{Incorporating Design Constraints.} Our continuous design space and sensor modeling enables rapid LiDAR system synthesis under arbitrary constraints. Spatial, angular, or temporal limits can be enforced by restricting the support of the implicit density $p(\mathbf d)$ to an admissible region $\mathcal C \subseteq \mathcal D$, i.e.\ setting $p(\mathbf d)=0$ for $\mathbf d\notin\mathcal C$, and refitting the sensor model $q(\mathbf d\mid\theta)$ to the truncated density. FoV and time‐gate bounds become simple bounds on the diagonal entries of $\Sigma_a$ and $\Sigma_\tau$ (e.g.\ $\Sigma_{a,ii}\in[\sigma_{\min}^2,\sigma_{\max}^2]$). The number of sensor components is controlled by the mixture order $G$, and any subset of parameters (e.g.\ $\mu_x,\Sigma_x$) can be fixed to enforce positional or motion constraints. We visualize examples of applying constraints in \cref{fig:fig3b_constraints}, demonstrating the flexibility of our technique for user-specified design conditions. 

\vspace{-2mm}
\section{Experiments} \label{sec:sec6}
\vspace{-4pt}
We evaluate our LiDAR design technique across three 3D vision tasks: face scanning for mesh reconstruction, robot arm end-effector tracking, and in-motion robotic warehouse object detection. In each setting, we compare our learned designs under varying scene and user-imposed constraints against: (i) uniform ray sampling over origin and angle with fixed time gating, and (ii) random sampling over origin and angle. Additional details, quantitative metrics, and qualitative results are provided in the supplement.

\begin{figure*}[tbh!]
\centering
\includegraphics[width=0.93\linewidth]{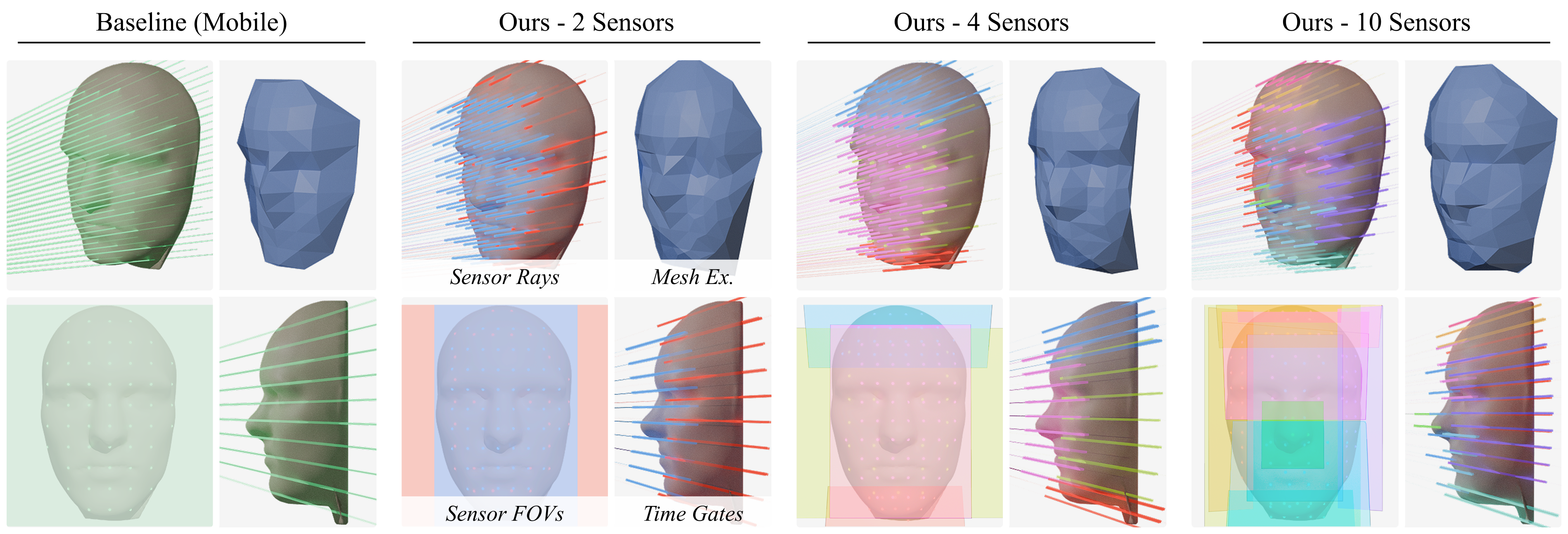}
\vspace{-3mm}
\captionof{figure}{\textbf{Automatically proposed sensor designs for face scanning.} We visualize ray-based sampling, FoVs, and time gates of learned face scanning sensor designs from our technique. Generated sensors can adapt to arbitrary user constraints and variations in facial geometry.\vspace{-2.5mm}}
\label{fig:figure4}
\vspace{-2pt}
\end{figure*}

\noindent \textbf{Expt. A: Face Scanning for Mesh Reconstruction.} We evaluate our method on smartphone flash-LiDAR design for face scanning. We sample 50 face meshes from the Basel Face Model \cite{paysan20093d} to learn an implicit LiDAR density; we then generate new designs with varying constraints over: (i) sensor count, which controls spatial design flexibility, and (ii) ray budget, the total number of sampled rays. For each proposed design, we simulate ray–mesh intersections on 50 test faces, reconstruct meshes via Delaunay triangulation, and evaluate accuracy using Chamfer distance \cite{barrow1977parametric}. As shown in \cref{fig:figure4}, the learned designs adapt to fine facial geometry, allocating coverage to key regions (e.g., a dedicated nose sensor in the 10-sensor design) and adjusting per-sensor sampling under fixed ray budgets. Our designs yield consistently higher reconstruction fidelity, and reduce bandwidth by 6$\times$ compared to uniform baselines (\cref{fig:results_metrics}).

\begin{figure*}[htb!]
\centering
\includegraphics[width=0.95\linewidth]{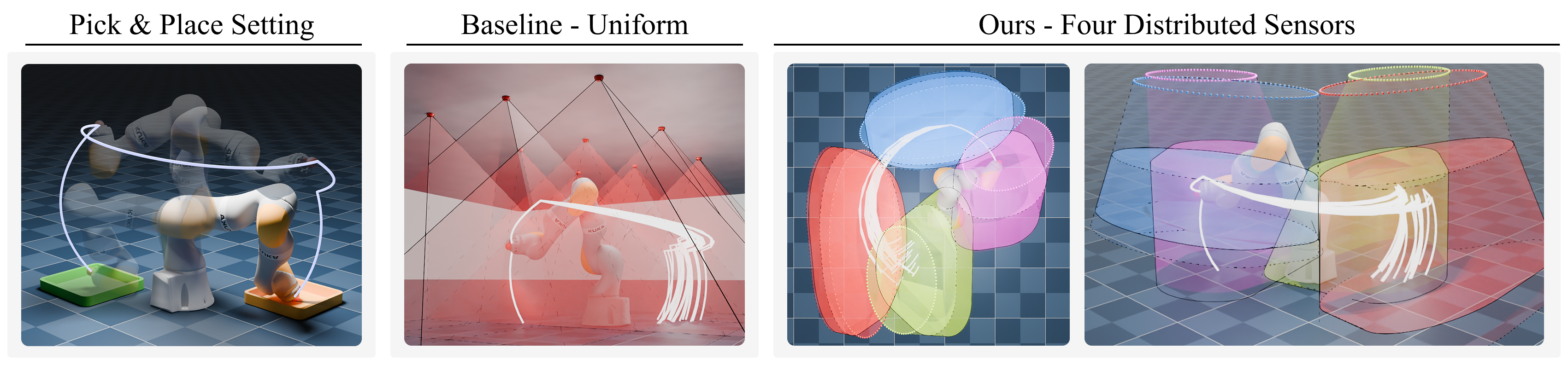}
% \framebox[\linewidth]{\rule{0pt}{2in}}
\vspace{-3mm}
\captionof{figure}{\textbf{Automatically proposed distributed LiDAR system for end-effector tracking}. Our technique enables distributed LiDAR system design, balancing sensor counts, placements, FoV, and time gating to enable robust tracking for arbitrary classes of scene motion.\vspace{-5mm}}
\label{fig:figure6}
% \vspace{-8pt}
\end{figure*}

\vspace{0.5mm}
\noindent \textbf{Expt. B: Robotic End Effector Tracking.} We evaluate our method for \textit{distributed LiDAR system} design in robotic end-effector tracking. Using MuJoCo~\cite{todorov2012mujoco}, we simulate a 7-DOF KUKA IIWA performing 40 randomized pick-and-place tasks with realistic approach, transit, and descent trajectories solved via inverse kinematics. Sensors are constrained to ceiling-mounted positions ($z=1$m), azimuth in $[0,2\pi]$ rad, elevation in $[-\pi/2,-\pi/3]$ rad, and max ToF 2m. We fit Gaussian sensors with learnable origin covariances over the implicit density, simulating distributed placements; we then importance-sample rays under a fixed budget, and simulate ray–end-effector intersections on held-out scenes. Reconstructed trajectories (via spline interpolation) are compared to ground truth trajectories using Fréchet distance ~\cite{alt1995computing}. As shown in \cref{fig:figure6}, our designs can accurately project \textit{scene motion variability} into the LiDAR design space, balancing spatial sampling, FoV, and time-gating. Across constraints, our designs achieve higher tracking accuracy with 2$\times$ lower bandwidth than uniform baselines.

\noindent \textbf{Expt. C: Adaptive Scanning for Object Detection.} We next evaluate \textit{motion-adaptive scanning LiDAR} design in a simulated warehouse inspection task. Across 100 scenes with randomized shelf heights and box sizes, we learn an implicit LiDAR density conditioned on robot motion ($x \in [0, X]$), with fixed azimuth $-\pi/2$, elevation in $[0, \pi/2]$, and maximum ToF 5\,m. At each robot position, we fit Gaussian-parameterized rays (with learned elevation and time gates) to the learned density, yielding motion-adaptive sensor configurations. Ray--mesh intersections are simulated on held-out scenes, with ray--box hits counted as detections. As shown in \cref{fig:teaser}, conditioning LiDAR design on robot location yields spatially adaptive sensing strategies; these designs achieve more reliable object detection while reducing bandwidth by $\sim$10$\times$ under a fixed ray budget (\cref{fig:results_metrics}).

\begin{figure}[t!]
  \centering
  \includegraphics[width=\columnwidth]{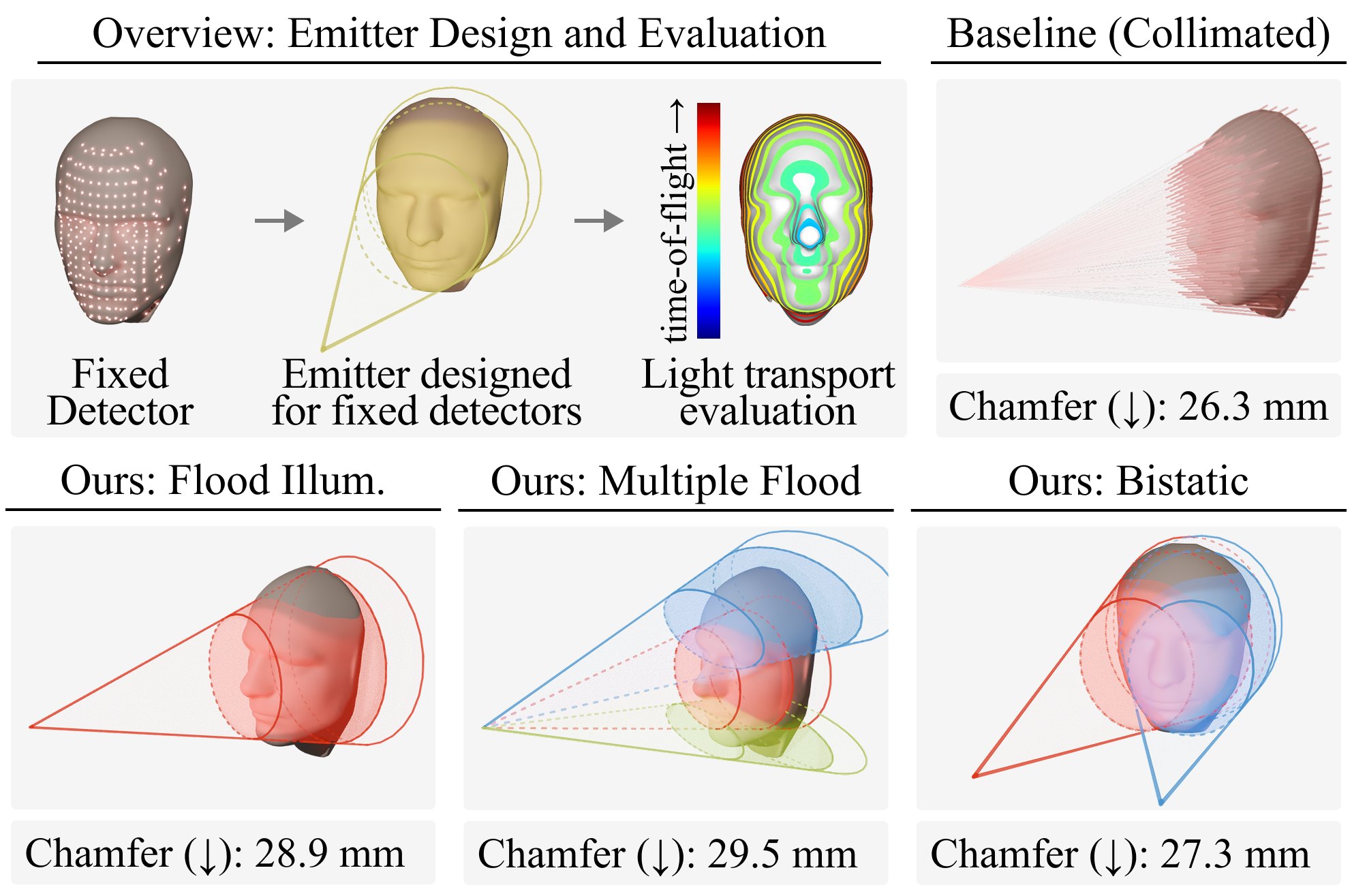}
  \vspace{-6mm}
  \captionof{figure}{\textbf{We enable diverse emitter design for LiDAR detectors} by conditioning the implicit density on a specific detector. Proposed emitters satisfy arbitrary placement and FoV constraints while maintaining robust task performance.}
  \label{fig:figure5}
  \vspace{-6mm}
\end{figure}

\begin{figure*}[t!]
\centering
\includegraphics[width=\linewidth]{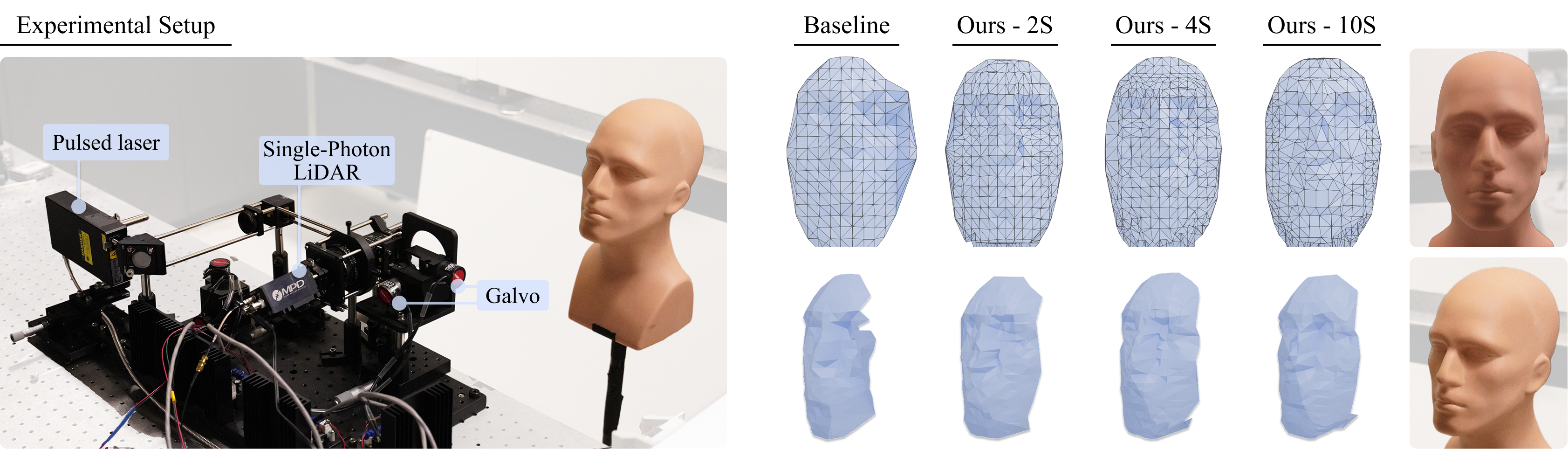}
\vspace{-8mm}
\captionof{figure}{\textbf{Real-world face scanning setup and qualitative results.} We validate our approach using a single-photon LiDAR that allows precise control of scanning angles and time gating, emulating our learned sensor configurations. The resulting reconstructions show improved surface coverage, finer detail, and robustness to depth outliers.}
\label{fig:figure8}
\vspace{-4mm}
\end{figure*}

\vspace{0.5mm}
\noindent \textbf{Expt. D: Emitter Design.} Our learned implicit density also supports automatic synthesis of constraint-aware emitters for a fixed LiDAR detector. In real-world LiDARs, limitations on emitter placement or power often prevent ideal co-located, collimated laser illumination. As shown in \cref{fig:figure5}, given a fixed detector replicating a standard smartphone LiDAR, our method can fit diverse emitter designs in 6D design space, yielding single- and multi-flood illumination patterns and bistatic designs under user-defined constraints. We evaluate these designs in a physically-accurate light-transport simulator \cite{pediredla2019ellipsoidal} and show that our alternative emitter designs yield reconstruction quality comparable to conventional co-located laser designs (\cref{fig:figure5}).

\begin{figure}[t!]
  % \vspace{-6pt}                
  \centering
  \includegraphics[width=\columnwidth]{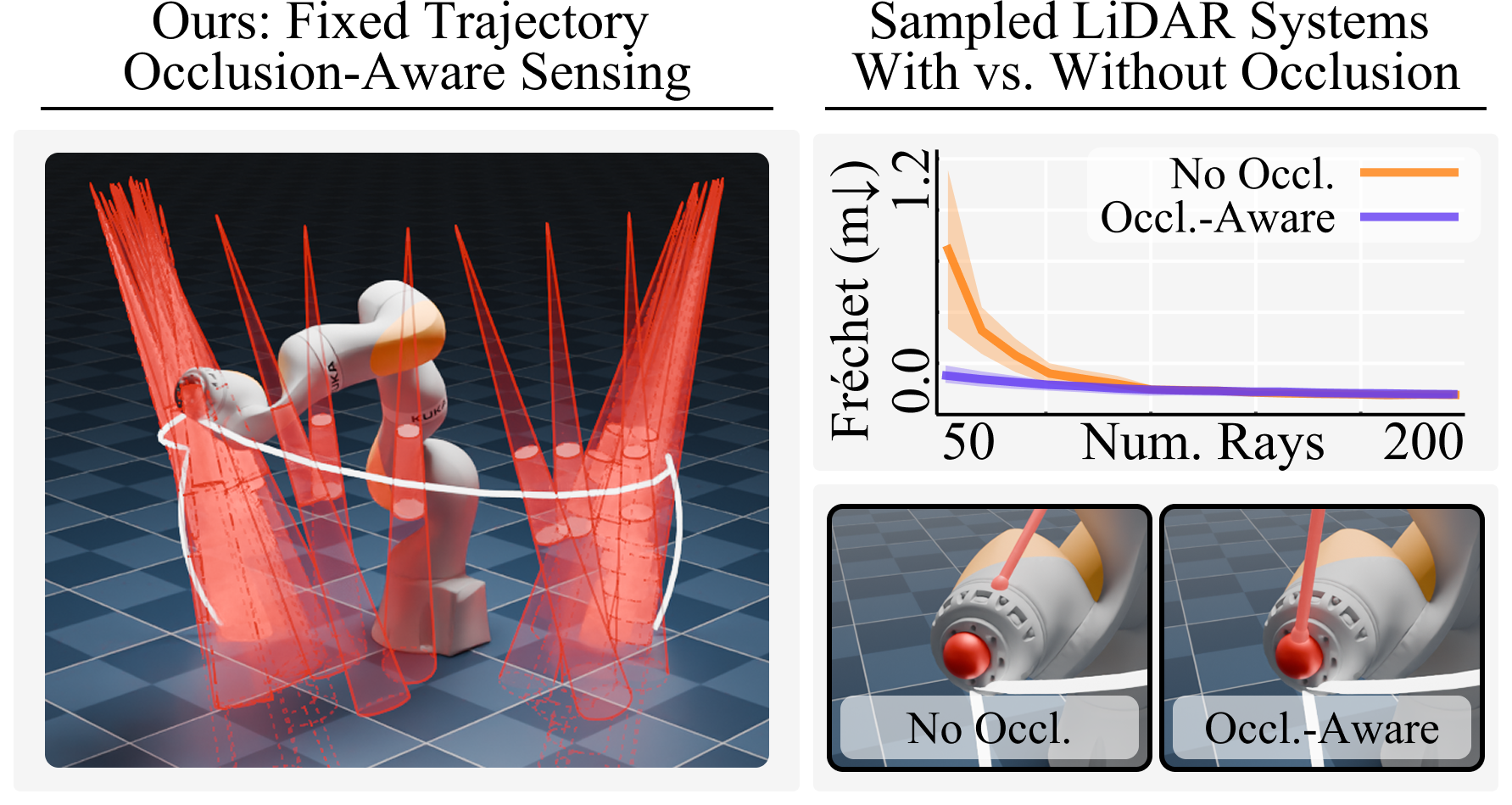}
  \vspace{-8mm}
  \captionof{figure}{\textbf{Occlusion-aware modeling improves LiDAR sampling.} Distributed systems generated from our occlusion-aware density produce higher-visibility ray samples, yielding better downstream performance than occlusion-naive designs.}
  \label{fig:figure7}
  \vspace{-6mm}                     
\end{figure}

\vspace{0.5mm}
\noindent \textbf{Expt. E: Ablations and Additional Baselines.} A key strength of our method is modeling scene occlusion effects through the ray-visibility term, which is critical for robust LiDAR design. To illustrate this, we compare our full density formulation against an ablated version without the ray-visibility term for end-effector tracking (\cref{fig:figure7}). We sample distributed LiDAR configurations from each density and evaluate trajectory reconstruction accuracy (Fréchet distance) under varying ray budgets. Our occlusion-aware designs achieve comparable accuracy with 2$\times$ \textit{fewer rays} than configurations derived from the occlusion-naive density. Additional baselines against occupancy-grid modeling (see supplement) highlight the importance of our continuous and visibility-aware implicit density.

\vspace{0.5mm}
\noindent \textbf{Expt. F: Real-World Qualitative Evaluation.} Our method uses task-specific scene simulations to design LiDAR systems robust to real-world conditions. While earlier results evaluated synthesized systems on held-out simulations, we now validate our face-scanning design (Expt. A) on physical hardware, shown in \cref{fig:figure8}. We implement our proposed LiDAR designs using a single-photon avalanche diode (Micro Photon Devices PDM series) and a pulsed laser emitter ($\approx$50 ps resolution) in a confocal setup with a two-axis galvanometer scanner ($\pm 20 ^\circ$ per axis $\approx40^\circ$ total FoV) to steer the pulsed beam along learned ray directions. We emulate our 2, 4, and 10-sensor configurations (each with 576 total rays) against a uniform scanning baseline typical of mobile devices. As shown in \cref{fig:figure8}, our learned designs yield denser surface coverage, recover finer facial details, and suppress noise-induced depth discontinuities. These results suggest that our learned LiDAR designs remain effective under real-world LiDAR conditions (e.g., noise, time jitter, non-ideal pulse). Additional real-world evaluation details are included in the supplement. 

\vspace{0.5mm}
\noindent \textbf{Limitations.}  \label{sec:sec6_1}
Our framework relies on accurate scene simulations to learn task-driven LiDAR designs; as such, our method's performance may degrade when simulations fail to capture real-world variability, or when test scenes fall outside the simulated distribution. While large-scale real-world validation remains an important next step, evaluation on existing pre-captured datasets (e.g., KITTI \cite{geiger2013vision}) is not directly feasible, since our flow-based model requires actively querying the continuous 6D LiDAR design space. Extending our technique to large-scale real-world deployments is an important direction for future research.

\vspace{-2mm}
\section{Conclusion} \label{sec:sec7}
\vspace{-2mm}
We present a framework for computational design of LiDAR systems that represents sensors in a continuous 6D design space. By learning task-driven implicit densities over this space, our method synthesizes novel, efficient LiDAR configurations for diverse 3D vision tasks, including reconstruction, tracking, and detection. We believe our work is a step towards a new paradigm in generative computational design -- enabling rapid development of unconventional yet efficient 3D vision systems for robotics, mobile devices, and autonomous systems.

%%%%%%%%%%%%%%%%%%%%%%%%%%%%%%%%%%%%%%%%%%%%%%%%%%%%%%%%%%%%

\paragraph{Acknowledgements.} Nikhil Behari is supported by the NASA Space Technology Graduate Research Opportunity Fellowship. Aaron Young is funded by the National Science Foundation (NSF) Graduate Research Fellowship Program (Grant \#2141064). Tzofi Klinghoffer is supported by the Department of Defense (DoD) National Defense Science and Engineering Graduate (NDSEG) Fellowship Program.

{
    \small
    \bibliographystyle{ieeenat_fullname}
    \bibliography{main}
}

\end{document}